\pdfoutput=1

\documentclass[11pt]{article}

\usepackage{EMNLP2023}
\usepackage{times}
\usepackage{latexsym}
\usepackage{multirow}
\usepackage{booktabs}
\usepackage{tabularx}
\usepackage{tcolorbox,lipsum}
\tcbuselibrary{listings, breakable, skins, theorems}
\usepackage{amsmath}

\usepackage{graphicx}

\usepackage[T1]{fontenc}

\usepackage[utf8]{inputenc}

\usepackage{microtype}
\usepackage{enumerate}
\usepackage{inconsolata}

\makeatletter
\newcommand*{\rom}[1]{\expandafter\@slowromancap\romannumeral #1@}
\makeatother

\newtcolorbox[auto counter, number within=section]
{promptbox}[2][]{
colback=gray!10!white,
colframe=gray!60!gray,
fonttitle=\bfseries\sffamily,
title=Prompt~\thetcbcounter: #2,
rounded corners,
arc=1.3mm,
boxrule=0.5pt,
enhanced,
breakable,
listing only,
listing options={
    basicstyle=\ttfamily\bfseries\itshape\fontsize{5}{6},
    numbers=left,
    numberstyle=\tiny\color{gray!80!black},
    stepnumber=1,
    numbersep=5pt,
    showspaces=false,
    showstringspaces=false
},
label={prompt:#1}
}

\title{Assessing ``Implicit'' Retrieval Robustness of Large Language Models}

\author{Xiaoyu Shen\textsuperscript{1} \,Rexhina Blloshmi\textsuperscript{2}\thanks{\;Work Done Outside Amazon} \, Dawei Zhu\textsuperscript{3} \, Jiahuan Pei\textsuperscript{4} \, Wei Zhang\textsuperscript{1}\thanks{\;Corresponding Author} \\
\textsuperscript{1}Eastern Institute of Technology, Ningbo\\
\textsuperscript{2}Amazon AGI, \textsuperscript{3}Saarland University, Saarland Informatics Campus\\
\textsuperscript{4}Centrum Wiskunde \& Informatica, Amsterdam, The Netherlands\\
\texttt{\{xyshen,zhw\}@eitech.edu.cn}}

\begin{document}
\maketitle
\begin{abstract}
Retrieval-augmented generation has gained popularity as a framework to enhance large language models with external knowledge. However, its effectiveness hinges on the retrieval robustness of the model. If the model lacks retrieval robustness, its performance is constrained by the accuracy of the retriever, resulting in significant compromises when the retrieved context is irrelevant. In this paper, we evaluate the ``implicit'' retrieval robustness of various large language models, instructing them to directly output the final answer without explicitly judging the relevance of the retrieved context. Our findings reveal that fine-tuning on a mix of gold and distracting context significantly enhances the model's robustness to retrieval inaccuracies, while still maintaining its ability to extract correct answers when retrieval is accurate. This suggests that large language models can implicitly handle relevant or irrelevant retrieved context by learning solely from the supervision of the final answer in an end-to-end manner. Introducing an additional process for explicit relevance judgment can be unnecessary and disrupts the end-to-end approach.~\footnote{We release our model outputs \href{https://drive.google.com/drive/folders/1Mrx2E0E2KaDiye_oeZ2wQuC4boKOisY7?usp=sharing}{here}. The used datasets can be accessed through \href{https://drive.google.com/drive/folders/1Pv1gWX3xiI2qs2pHddQHqn91aSZK1d1w?usp=sharing}{this link}.}
\end{abstract}

\section{Introduction}
Large language models (LLMs) have brought about a paradigm shift in the field of Natural Language Processing, enabling remarkable advancements in various tasks~\cite{brown2020language,su2022rocbert,su2022welm,chowdhery2023palm,achiam2023gpt}. However, their static nature imposes limitations, preventing them from fully encompassing all specialized knowledge or maintaining its currency~\cite{dhingra2022time,kandpal2023large}. To mitigate this constraint, a prevailing trend involves the adoption of retrieval-augmented generation (RAG) methodologies~\cite{guu2020retrieval,lewis2020retrieval,izacard2022few}. Through bringing extra context from the retriever, these models can tap into external knowledge reservoirs, refining their outputs with heightened precision and contextually fitting information~\cite{wang2023survey,gao2023retrieval,chen2023combating}. 

\begin{figure}[t]
    \centering
    \includegraphics[width=1\columnwidth]{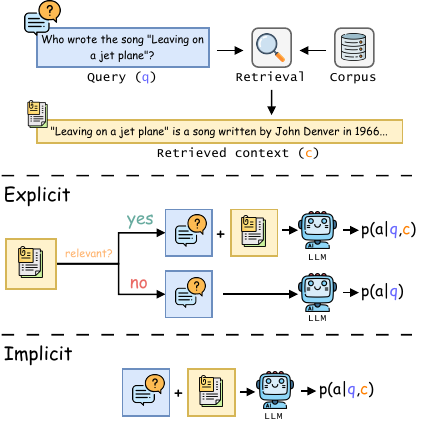}
    \caption{\small Difference between explicitly and implicitly modelling the relevance of retrieved context. The explicit approach evaluates whether the retrieved context is relevant and then calls different functions based on this assessment. In contrast, the implicit approach directly generates the final answer in an end-to-end manner.}
    \label{fig:idea_illustration}
\end{figure}

Nevertheless, acquiring a reliable retriever is challenging. Since the number of candidate documents for retrieval is typically much larger than the vocabulary size of LLMs, it is often easier to generate the correct answer from the knowledge stored in the model parameters rather than retrieving it~\cite{yu2023generate,maekawa2024retrieval,Feldman2024ragg}. When the retriever is imperfect, the quality of LLM generations can be significantly compromised, which often leads to poorer performance compared to scenarios where no retriever is employed at all~\cite{li2022large,luo2023sail}.

The main reason that influences the quality of RAG is their \emph{retrieval robustness}~\cite{yoran2024making}. Ideally,
a retrieval-robust model should possess two key capabilities:
\begin{enumerate}[I]
    \item Properly incorporate helpful retrieved information to provide an accurate answer.
    \item Ignore distracting information and rely on its own internal knowledge as a fallback.\footnote{Some works take a conservative strategy of refraining from answering if the retrieved context is unhelpful. However, this limits the model's potential to the accuracy of the retriever and underutilizes LLMs' internal knowledge~\cite{li2023trac}.}
\end{enumerate}
Capability \rom{1} pertains to scenarios where the retrieved information aids in deriving the answer, while Capability \rom{2} pertains to scenarios where the retriever only returns distracting information.

A wide range of approaches have been proposed to improve the retrieval robustness of LLMs, which can be classified into two categories: The first category \emph{explicitly} decouples Capability \rom{1} and \rom{2} by injecting an intermediate process to judge the relevance of retrieved information, based on which different functions are called~\cite{creswell2022faithful,yu2023chain}. The second category, on the contrary, relies on the model itself to \emph{implicitly} judge the relevance of the retrieved information and generate the right answer directly~\cite{luo2023sail,yoran2024making}. Figure~\ref{fig:idea_illustration} depicts the difference between explicit and implict approaches.

Despite being finer-grained, explicit approaches increase runtime latency and the risk of error propagation. They also require annotations regarding the relevance of retrieved information, which can be costly to obtain on a large scale.\footnote{Annotations can be circumvented by developing complex self-supervision or weak-supervision algorithms~\cite{wang2024self}, but these algorithms often come with additional costs, such as increased computations or suboptimal performance.} In this paper, we conduct a thorough analysis in a controlled setting to evaluate the ``implicit'' retrieval robustness of LLMs. More concretely, we aim to determine \emph{the extent to which we can uphold the retrieval robustness without requiring explicit judgment of the retrieval's relevance}.

To conduct this analysis, we run extensive experiments with 5 question-answering tasks spanning different domains and scenarios; 5 open-source LLMs (Vicuna-7/13/33B and Llama 2-7/13B); 2 closed-source models (GPT-3.5 and GPT-4) and 3 testing scenarios (zero-shot with prompting, full fine-tuning and LoRA fine-tuning). For each experiment, we run controlled tests to evaluate Capability \rom{1} and \rom{2} of the models separately. Our findings can be summarized as follows:
\begin{itemize}
    \item Without fine-tuning, open-source LLMs often under-perform GPT-3.5/GPT-4 in terms of Capability \rom{1}, but match them in terms of Capability \rom{2}. Larger models generally exhibit greater resilience to distractions.
    \item Fine-tuning on gold context enhances Capability \rom{1} on challenging tasks, but often hits a plateau on easier tasks, accompanied by a drop in Capability \rom{2}. LoRA matches full fine-tuning in improving Capability \rom{1} and better preserves Capability \rom{2}.
    \item Fine-tuning on noisy context can significantly enhance Capability \rom{2} of LLMs \emph{without} affecting their Capability \rom{1}. A higher noise ratio (50\%) can often lift the performance of Capability \rom{2} to the level of non-retrieval models, except on questions requiring multi-hop or multi-turn inference.
\end{itemize}
Overall, we suggest that LLMs are notably robust at noisy retrievals during fine-tuning. With a high noise ratio, the ``implicit'' retrieval robustness of LLMs can be remarkably effective. For most question-answering tasks that do not involve sophisticated multi-hop or multi-turn inference, relying on the model's implicit retrieval robustness may already suffice.
\section{Related Work}
\paragraph{Retrieval-Augmented Generation}
Due to the static nature of the knowledge stored within their parameters, large language models encounter difficulties in tasks that require extensive knowledge or have temporal dependencies~\cite{qiu2023large}. Retrieval-augmented generation has emerged as a valuable approach to address these limitations by enabling models to retrieve and integrate information from external sources during the generation process~\cite{guu2020retrieval,lewis2020retrieval,del2021question,del2022rewriting}. The external sources may include knowledge bases, search engines, multi-turn histories, or private databases, depending on the specific knowledge needed for the task~\cite{gao2023retrieval}. Various studies have explored the integration of retrieval mechanisms into generative models to enhance the quality and relevance of generated text from LLMs~\cite{peng2023check,shi2023replug,ren2023investigating}. The retrieval-augmented mechanism not only improves performance but also offers a cost-effective approach to adapting the model for diverse domains by dynamically adjusting external knowledge sources~\cite{barlacchi2022focusqa,ram2023context}. Although improvement has been observed, the quality of generations is strongly affected by the accuracy of retrievers. Inaccuracies in retrievers can lead to the incorporation of irrelevant or misleading information, resulting in lower-quality generated content~\cite{xu2024list,Feldman2024ragg}.

\paragraph{Retrieval-Robust Large Language Model}
Recognizing that the quality of text generations from LLMs is significantly influenced by the retriever's quality, various research works have been proposed to enhance the retrieval robustness of LLMs, i.e.\ , the model should effectively utilize accurate retrieved information while also disregarding distracting information in cases where the retriever is inaccurate~\cite{yoran2024making}. The first line of research introduces an intermediary step to assess the relevance of retrieved information, aligning with conventional methods of step-by-step planning in text generation~\cite{konstas2013global,moryossef2019step,shen2020neural}. When the information is detected to be unhelpful, the model will simply fall back to use its own parameterized knowledge to answer the question. This helpfulness label is usually obtained by manual annotation~\cite{glaese2022improving,shuster2022blenderbot}, chain-of-thought prompting on a powerful LLM~\cite{creswell2022faithful,yu2023chain,zhang2024adapt}, or inspecting its effect on the model generation~\cite{Jeong2024adaptive}. Although this step-by-step approach provides finer-grained signals, it also leads to increased runtime latency and training costs, with potential risks of error propagation~\cite{wang2023survey}. Conversely, the alternative line of research employs an end-to-end approach to train models to autonomously discern the relevance of retrieved information from without extra helpfulness labels. The key to achieving successful end-to-end learning is to incorporate noisy retrievals, allowing the model to adjust to distracting information~\cite{luo2023sail,yoran2024making}. Nonetheless, existing studies lack quantitative analysis on how the retrieval robustness is influenced by factors such as the model, fine-tuning method, data, and noise ratio. Our research seeks to address this gap in the literature.

\section{Definition of Retrieval Robustness}
Let $q,c,a$ denote the question, context retrieved from an external source, and answer respectively. The variable 
$p$ denotes the probability estimator from the LLM generator. In retrieval-augmented generation, the retriever retrieves some context $c$\footnote{Depending on the granularity of the retrieval, the context can be in the unit of documents, passages, sentences, entities, etc~\cite{shen2022low}.} from external sources where $c$ can be either helpful or unhelpful depending on the accuracy of the retriever. The answer is generated from $p(a|q,c)$ by conditioning on $q$ and $c$. An LLM is considered \emph{retrieval-robust} if the probability estimation $p(a|q,c)$ remains effective regardless of the helpfulness of $c$. It corresponds to two different capabilities that the LLM should possess:
\begin{enumerate}[I]
    \item When $c$ is helpful, i.e.\ , the correct answer $a^*$ can be derived from the information contained in $c$, then it should return $a^*$.
    \item When $c$ is not helpful, it should discard the information in $c$ and rely on its own parameterized knowledge $p(a|q)$ to answer the question.
\end{enumerate}
Equation~\ref{eq:two_capabilities} illustrates the ideal $p_\text{robust}(a|q,c)$ from a retrieval-robust LLM mathematically, where $\delta$ is the dirac-delta function.
\begin{equation}
    p_\text{robust}(a|q,c)= 
\begin{cases}
    \delta({a-a^*}),& \text{if } a^* \in c\\
    p(a|q),              & \text{otherwise}
\end{cases}
\label{eq:two_capabilities}
\end{equation}
\section{Experiment Setup}
\paragraph{Model}
We test 5 open-source LLMs: Vicuna-1.3-7/13/33B~\cite{vicuna2023} and Llama 2-chat-7B/13B~\cite{touvron2023llama}, as well as two closed-source LLMs GPT-3.5 and GPT-4~\cite{achiam2023gpt}. For open-source LLMs, we test their performance with zero-shot prompting, LoRA and full fine-tuning on task-specific datasets. For closed-source LLMs, we only report their performance by prompting them with instructions.

\paragraph{Dataset}
In order to test model capabilities comprehensively, we test the models on 5 datasets covering diverse domains, question types and knowledge sources: AmbigQA~\cite{min2020ambigqa}, ePQA~\cite{shen2022product,shen2022semipqa}, Musique~\cite{trivedi2022musique}, SciQ~\cite{welbl2017crowdsourcing} and TopioCQA~\cite{adlakha2022topiocqa}. 

We specifically choose datasets with short answers because evaluating long answers is known to be challenging~\cite{xu2023critical}.
AmbigQA is a refined version of Natural Questions~\cite{kwiatkowski2019natural} after removing the ambiguity among questions. It contains general-knowledge questions answerable with Wikipedia contents.
ePQA contains product-specific questions from the Amazon website. Testing on ePQA reduces the chance that the model
memorizes the knowledge since product information is tail-distributed.
MuSiQue is an improved version of HotpotQA~\cite{yang2018hotpotqa} after removing potential short cuts. It contains questions requiring multi-hop reasoning, which have to be answered with at least two passages.
SciQ contains scientific questions about physics, chemistry, etc. TopioCQA contains questions in multi-turn conversations. Table~\ref{tab:dataset} provides a summary of used datasets. Dataset examples are in Appendix~\ref{sec:app_dataset}.

\begin{table}
\centering
\small
\begin{tabular}{ccc}
\toprule
\textbf{Dataset} & \textbf{Question} & \textbf{Knowledge Source} \\
\midrule
AmbigQA & General-Knowledge & Wikipedia\\
ePQA & Product-Specific & Amazon\\
Musique & Multi-Hop & Wikipedia\\
SciQ & Scientific & TextBook\\
TopioCQA & Conversational & Wikipedia\\
\bottomrule
\end{tabular}
\caption{\label{tab:dataset}
\small Datasets used in this paper. We choose 5 datasets with diverse question types and knowledge sources.
}
\end{table}

\paragraph{Hyperparameter} When fine-tuning models, we observe that the learning rate can have big impact on the performance. In general for 7B/13B models, full fine-tuning
requires a small learning rate (in the scale of 1e-6) while LoRA fine-tuning requires a larger learning rate (in the scale of
1e-4). For 33B models, a small learning rate in the scale of 1e-6 is necessary. Due to the large impact of learning rate, we perform a grid search over [1e-6, 3e-6, 5e-6, 1e-5, 3e-5, 5e-5, 1e-4, 3e-5, 5e-4, 1e-3, 3e-3,
5e-3] for every model fine-tuning in the following section, then choose the checkpoint with the best score.\footnote{As the learning rate increases, the behavior of the curve varies between full FT and LoRA FT. In full FT, the model performance initially improves before declining. The optimal rate falls somewhere in between. In LoRA FT, the model performance fluctuates, showing two cycles of improvement and decline, with the optimal rate located at one of the peaks.} The batch size is fixed as 64 for all runs.
The model is fine-tuned for 1 epoch with the best-performing learning rate. 

\paragraph{Prompt} We conduct a series of prompt engineering and finalize two prompt templates: Template~\ref{prompt:woret} is used when the retrieval is not involved and~\ref{prompt:wret} is used when the retrieval is involved. For the ePQA dataset, we add an additional instruction to let the model always start with ``yes/no'' for binary questions to enable easier evaluation. For the TopiOCQA dataset, we further instruct the LLM to be aware that the question is within a conversation and turns are separated by the <SEP> symbol. Details are in Appendix~\ref{sec:app_prompt}. Empirically we find these templates are the best at inducing LLMs to produce answers at the desired format. In order to keep a fair comparison, we use the same set of prompts both when directly prompting the original LLMs, and when fine-tuning them, such that we can quantify how fine-tuning changes the retrieval robustness.

\begin{promptbox}[woret]{Instruction w/o Retrieval}
  \small \texttt{Answer the following question with less than 10 words. Question: }
  \textbf{[Q]}
\end{promptbox}

\begin{promptbox}[wret]{Instruction w. Retrieval}
  \small \texttt{Answer the following question with less than 10 words. The context is retrieved information which may or may not be helpful. When the context is unhelpful, answer it with your own knowledge. Question: }\textbf{[Q]} \texttt{Context: }\textbf{[C]}
\end{promptbox}

\paragraph{Metric} We evaluate the model's performance using recall, which indicates the number of words (excluding punctuation) from the gold answer that also appear in the model prediction. The recall metric is averaged across the test samples. This choice is made because LLMs may generate answers that are correct but longer than the concise answers in the original dataset, so using other metrics such as precision or F1 scores can significantly underestimate their performance~\cite{adlakha2023evaluating}. Empirically we also observe that the recall score correlates the best with human evaluations by manually examining 100 cases from each dataset.

\paragraph{Evaluation} We evaluate the model performance under three scenarios to quantitatively measure the two capabilities of retrieval robustness: (1) when no retrieval is provided; (2) when gold retrieval is provided; and (3) when distracting retrieval is provided. The gold retrieved information is extracted from the original dataset. To acquire the distracting retrieval, we retrieve the top 10 documents from the knowledge sources of each dataset.\footnote{We adopt a dense passage retriever~\cite[DPR]{karpukhin2020dense} trained on each knowledge source.} Subsequently, we consider the document with the lowest recall score with the answer as distracting information.\footnote{Most passages selected by this way have a recall score of 0 and only $\sim2\%$ of them have recall scores > 0.5, so we can consider they are almost distracting information.} The rationale for selecting from the top-10 DPR results is to align the process with realistic use cases. If the passages are blatantly distracting, it could make it too simplistic for the model to differentiate. We run all model generations with beam search under the beam size of 5.
\section{Results and Analysis}
We evaluate how retrieval robust different LLMs are in three scenatios: when directly prompting the original LLMs without fine-tuning them; when fine-tuning them only on gold context, and when fine-tuning them on mixed gold and distracting context. The results are presented in this order. Full results tables are in Appendix~\ref{sec:app_result}
\subsection{Without Fine-Tuning}
\begin{figure}[ht]
    \centering
    \includegraphics[width=1\columnwidth]{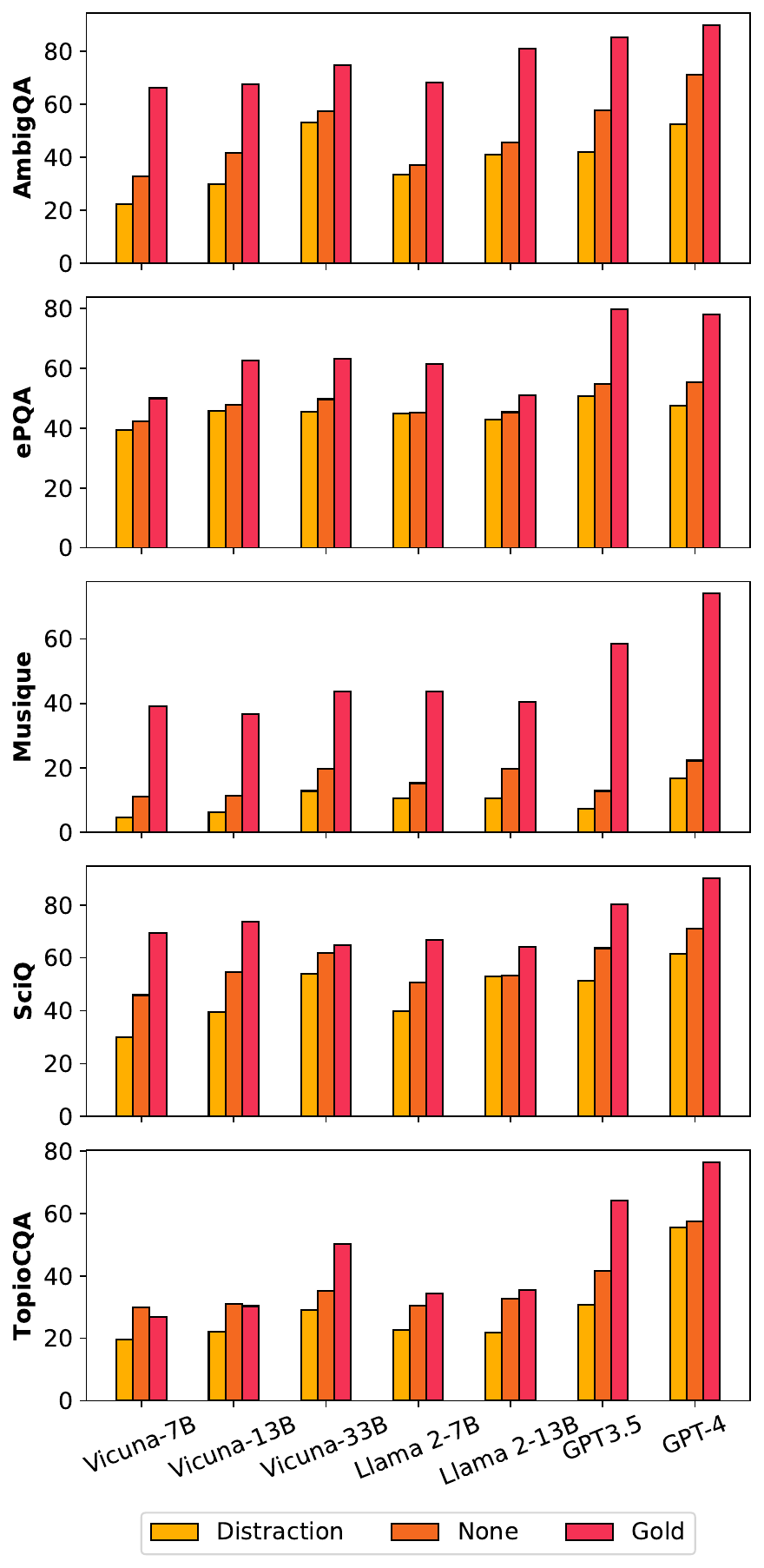}
    \caption{\small Performance by Prompting different LLMs when provided with no context (None), gold context (Gold) and distracting context (Distract).}
    \label{fig:woft}
\end{figure}
Figure~\ref{fig:woft} presents the results of directly prompting original LLMs without fine-tuning when provided with (1) no context, (2) gold context and (3) distracting context.
\paragraph{Without Context}
When no context is provided, LLMs often struggle to recall exact answers from their internal knowledge. As expected, larger models generally perform better than smaller ones. While GPT-3.5 and GPT-4 outperform open-source LLMs, their advantage is not substantial. For questions involving tail product knowledge (ePQA) or requiring multi-hop inferences (Musique), GPT-3.5 and GPT-4 face the same challenges as open-source models, limiting their advantages. Notably, most questions in ePQA are binary, allowing models to achieve decent scores through random guessing. As a result, performance on ePQA appears reasonable despite the LLMs' lack of specific product knowledge.
\paragraph{Capability \rom{1}}
When gold context is provided, all LLMs exhibit large improvement across all tasks, demonstrating their remarkable capabilities in extracting the right answers from the retrieved context. As model size increases, Vicuna-series models show more consistent performance improvements. However, for Llama 2-series models, the 13B model does not exhibit a clear advantage over the 7B model, except on the easiest dataset, AmbigQA. Nevertheless, \emph{there is still a large gap between open-source LLMs and closed-source GPT-3.5/4}. This gap is more notable ($>14\%$) on ePQA, Musique and TopioCQA as their question types and knowledge sources are more challenging. On ePQA, where a substantial amount of context is in JSON format, open-source LLMs encounter difficulty in efficiently processing information from this source. On Musique and TopioCQA, the presence of multiple items in the context and questions requires LLMs to accurately grasp the inter-dependencies among them, thereby increasing the complexity of the task.
\paragraph{Capability \rom{2}}
When distracting context is introduced, all LLMs experience a decline in performance compared to having no context at all. However, the decline with distracting context is usually much smaller than the gain from gold context, suggesting that \emph{existing LLMs are quite good at ignoring distracting context}.\footnote{Previous research typically reports larger declines because they did not explicitly instruct the LLM to revert to its own knowledge when the context is unhelpful~\cite{yoran2024making}.} The decline also varies across datasets. On datasets with tail knowledge, such as ePQA, the decline is minimal because the original LLM has almost no prior knowledge about specific products. Compared to Capability \rom{1}, there is a more consistent trend that larger models are more resilient with distracting context, suggesting that model size has a greater impact on the inherent capability for instruction following than on the understanding of additional context information. Surprisingly, \emph{powerful closed-source LLMs are even more vulnerable to distracting context, particularly on questions involving common knowledge} (AmbigQA and SciQ). The largest open-source LLM we tested, Vicuna-33B, is comparable to or better than GPT-3.5/4 in terms of performance drop when faced with distracting context.

In summary, when directly prompting LLMs, we have the following observations:
\begin{enumerate}
    \item In terms of Capability \rom{1}, open-source LLMs significantly under-performs GPT-3.5/4, especially on challenging tasks with complex question types and knowledge sources.
    \item In terms of Capability \rom{2},  open-source LLMs can be comparable or better than GPT-3.5/4. Larger models are more resilient with distracting context.
\end{enumerate}
\begin{figure}[ht]
    \centering
    \includegraphics[width=1\columnwidth]{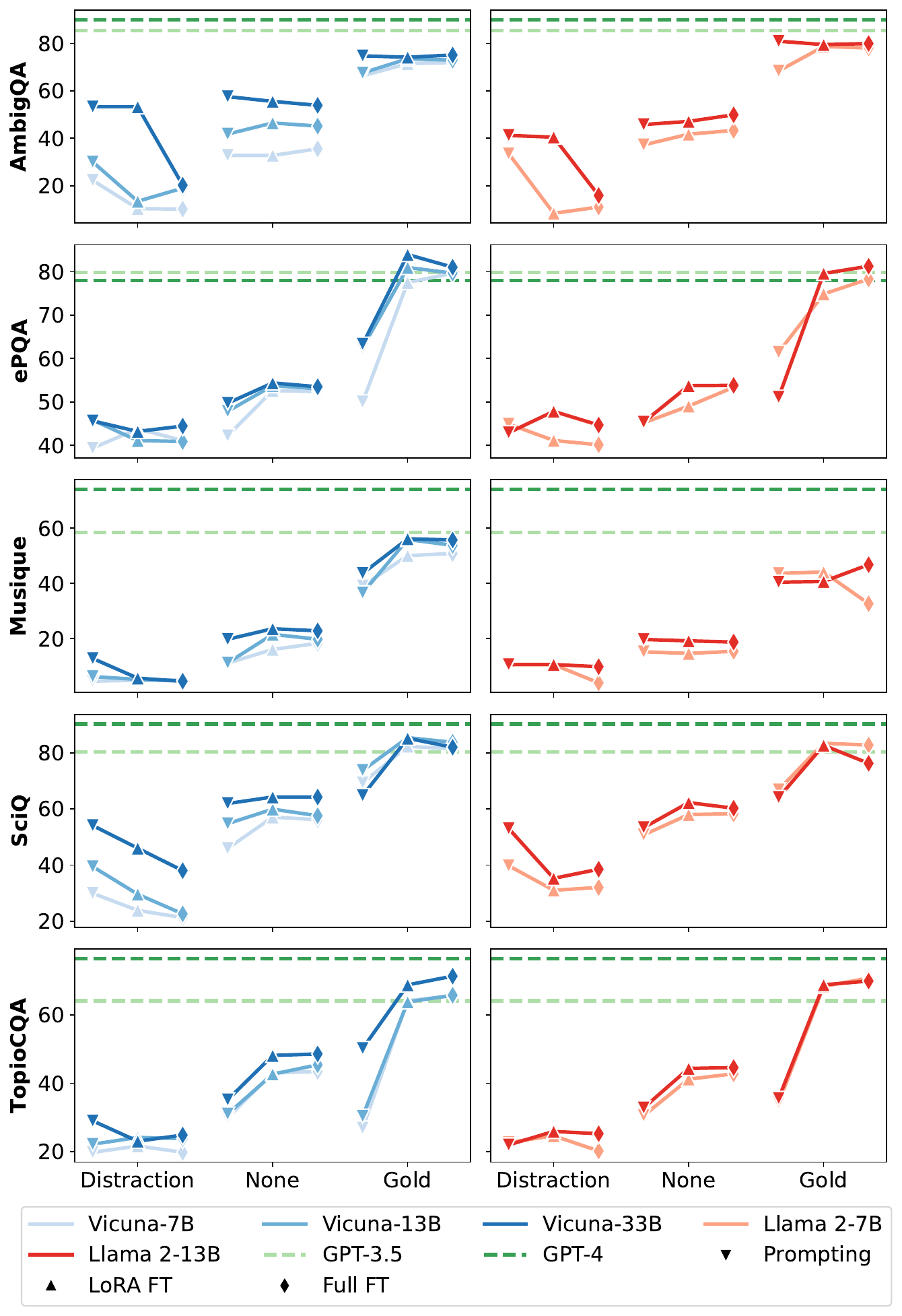}
    \caption{\small Performance by fine-tuning LLMs with and without context. When fine-tuning without context, we also test without context (None). When fine-tuning with context, we use only gold context when fine-tuning, then testing on gold and distracting context (Gold and Distraction).}
    \label{fig:gold_context}
\end{figure}
\subsection{Fine-Tuning on Gold Context}
While directly prompting existing LLMs can showcase remarkable performance, further task-specific fine-tuning is often necessary to fully tailor an LLM for a specific task. In order to see how task-specific fine-tuning can improve Capability \rom{1} and \rom{2} of LLMs, we perform full and LoRA fine-tuning on every task. During fine-tuning, the gold context is provided to teach LLMs to extract answers from the context, a common setup in retrieval-augmented training. Figure~\ref{fig:gold_context} depicts the experiment results.

\paragraph{Without Context}
Before fine-tuning on gold context, we first analyze the performance change when fine-tuning without context (``None'' as in Figure~\ref{fig:gold_context}). This can serve as an upper-bound performance for an LLM when the retrieved context is distracting ($p(a|q)$ as in Equation~\ref{eq:two_capabilities}). As observed, \emph{fine-tuning without context often results in limited improvement}. The only exception is the TopioCQA dataset, likely because the original LLMs struggle to understand the conversational format of the input and require fine-tuning to fully grasp the task format. This supports the superficial alignment hypothesis, which suggests that fine-tuning mainly trains the model to follow task-specific formats rather than adding new knowledge~\cite{zhou2024lima}.
\paragraph{Capability \rom{1}}
When fine-tuning LLMs with gold context, performance often improves significantly in terms of extracting the correct answer from the provided context. The improvement is especially pronounced on the ePQA and TopioCQA datasets, as these tasks are not inherently difficult but require adaptation to specific knowledge sources and conversational questions. On the ePQA dataset, the fine-tuned models can even outperform the closed-source GPT-3.5 and GPT-4 models.
After fine-tuning, there is a more consistent trend of larger models performing better, as the variance from prompting formats is reduced. However, all open-source LLMs struggle to further improve on the AmbigQA dataset, even with task-specific fine-tuning, possibly because their initial performance is already high and adding more data alone does not yield significant improvement.
Llama 2 models also hit a performance plateau on the Musique dataset. This suggests that \emph{task-specific fine-tuning alone may not be sufficient for open-source LLMs to match GPT-3.5 and GPT-4 in Capability \rom{1}}. Additional factors beyond task-specific fine-tuning might be necessary to close this gap. Across all models and datasets, there is no clear advantage of full fine-tuning over LoRA fine-tuning, even though training costs associated with full fine-tuning are significantly higher. 
\paragraph{Capability \rom{2}}
Despite the improvement of Capability \rom{1}, fine-tuning LLMs only on gold context can mislead them to always rely on the provided context, even when the information is distracting. This can eventually harm Capability \rom{2}, preventing LLMs from safely falling back to their internal knowledge. As observed in Figure~\ref{fig:gold_context}, there is indeed some performance decrease when LLMs are provided with distracting context. The gap between the LLM's probability estimation $p(a|q,c)$ and the ideal upper bound $p(a|q)$ widens. However, unexpectedly, the decrease is often small compared to the big performance boost when provided with gold context, especially on the more challenging ePQA, Musique and TopioCQA datasets. This may be because existing open-source LLMs struggle to handle distracting context on these more difficult datasets, so their initial performance is already close to random, leaving little room for further decline even when fine-tuning only on gold context. On the easier AmbigQA and SciQ datasets, LoRA fine-tuning often results in less performance drop compared to full fine-tuning due to the smaller number of adjustable training parameters.

In summary, when fine-tuning LLMs only on gold context, we have the following observations:
\begin{enumerate}
    \item Capability \rom{1} is improved significantly on challenging datasets, but hit a plateau on easier ones, suggesting other factors might be needed to fully close the gap with GPT-3.5/4.
    \item Capability \rom{2} is decreased mainly on easier datasets, potentially because the original performance on harder datasets with distracting context is already close to random.
    \item LoRA fine-tuning is similar to full fine-tuning in terms of improving Capability \rom{1}, but better at maintaining capability \rom{2}.
\end{enumerate}
\subsection{Fine-Tuning on Mixed Context}
\label{sec:mix_ret}

\begin{figure}[ht]
    \centering
    \includegraphics[width=1\columnwidth]{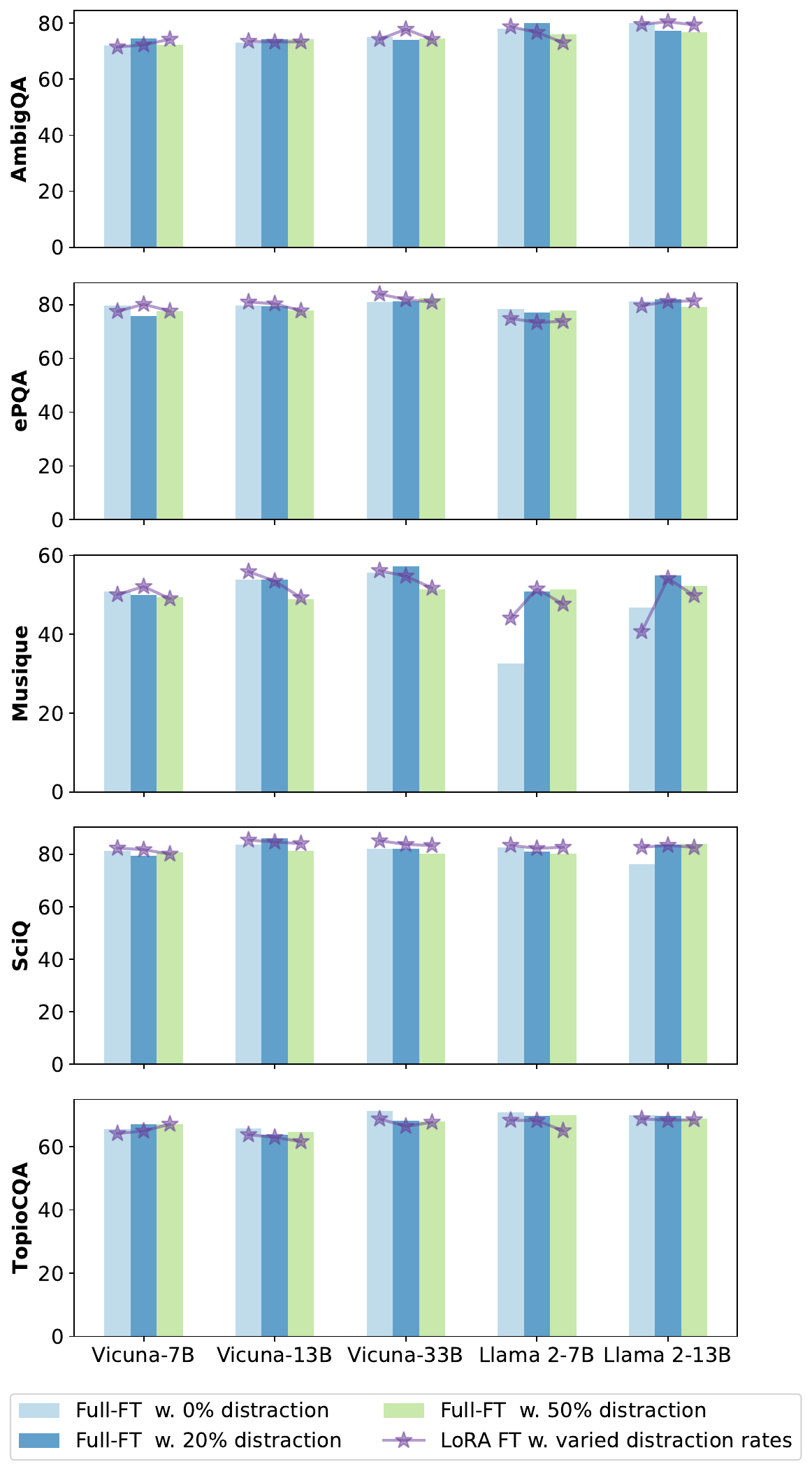}
    \caption{\small Fine-tuning LLMs with varied distraction ratios and then testing on gold context. Incorporating distracting context during fine-tuning does not compromise performance when provided with gold context.}
    \label{fig:mixed_ft_gold_test}
\end{figure}

\begin{figure}[ht]
    \centering
    \includegraphics[width=1\columnwidth]{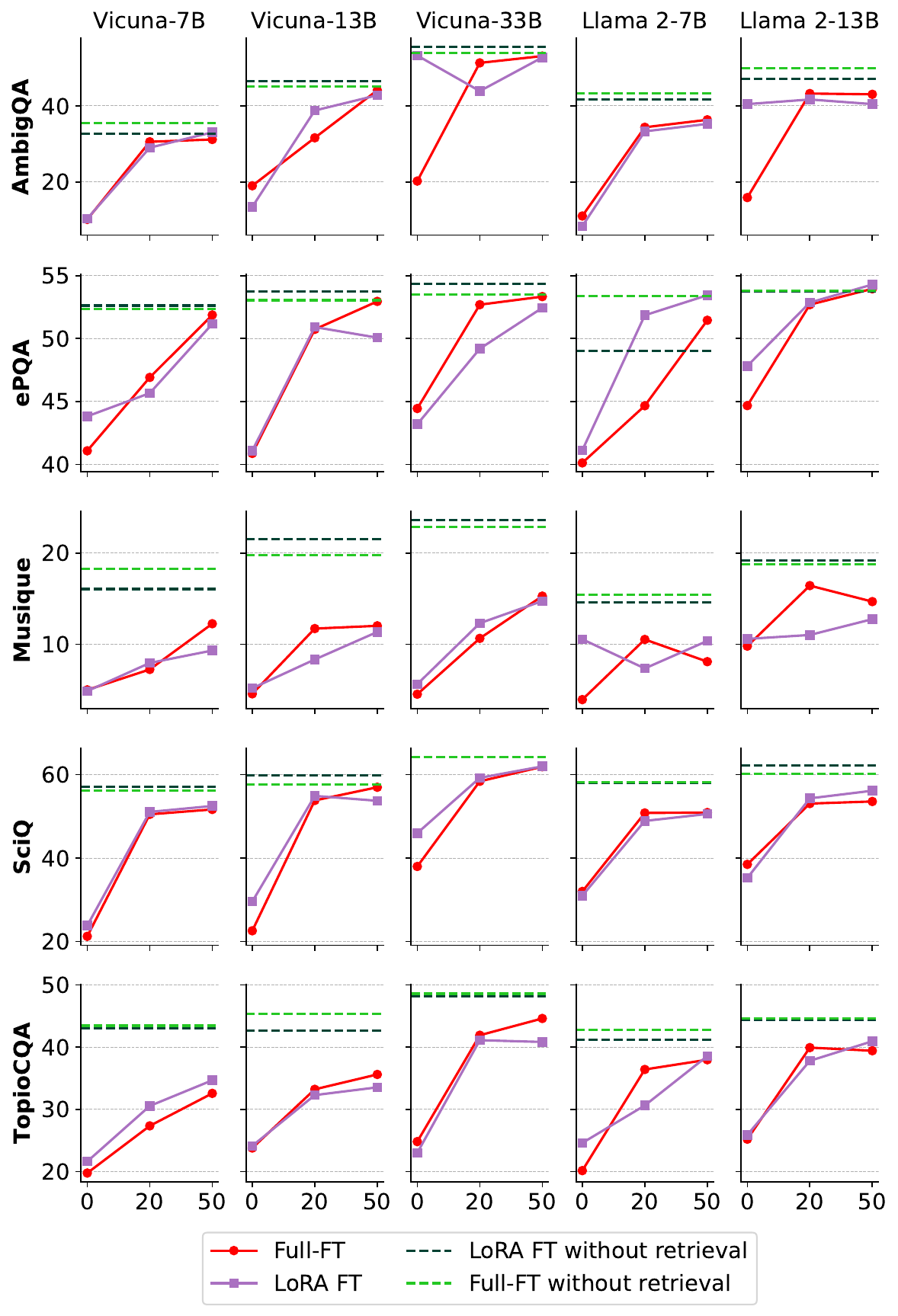}
    \caption{\small Fine-tuning LLMs with varying distraction ratios (0\%, 20\% and 50\%) and then testing on distracting contexts. Incorporating distracting context during fine-tuning significantly enhances retrieval robustness in distracting contexts. When the distraction ratio is increased to 50\%, LLMs can achieve performance comparable to the upper-bound performance without retrieval.}
    \label{fig:mixed_ft_distract_test}
\end{figure}

Fine-tuning LLMs solely with gold context can reduce their robustness to distracting context, which are inevitable in real-world retrieval-augmented generation scenarios. Therefore, we further explore whether the retrieval robustness can be improved by mixing distracting context into the fine-tuning datasets. We experiment with two distraction ratios: 20\% and 50\%. All distracting context are hard negative samples from the top-10 retrieved contents with dense retrieval to simulate real-case scenarios.
\paragraph{Capability \rom{1}}
Figure~\ref{fig:mixed_ft_gold_test} illustrates the performance of LLMs when fine-tuning with varying distraction ratios and testing on gold context. The results indicate that \emph{different levels of distracting context have little impact on performance}. Even when fine-tuned with 50\% distracting context (i.e.\, the training examples with gold context is reduced to half), the models still maintain their performance on gold context. Interestingly, in several instances, especially on challenging datasets such as Musique, augmenting the fine-tuning datasets with more distracting context actually enhances performance on gold context. This suggests that \emph{Capabilities \rom{1} and \rom{2} are not mutually exclusive}, and that incorporating some noisy context during fine-tuning can also be advantageous for Capability \rom{1}. Regarding the fine-tuning methods, LoRA fine-tuning performs similarly to full fine-tuning, with the only exception being observed on the Musique dataset for the Llama 2-7B model. This is due to the fact that fine-tuning cannot further enhance performance, allowing LoRA to preserve the original model performance to the greatest extent possible.
\paragraph{Capability \rom{2}}
After confirming that mixing distracting context into the fine-tuning dataset will not affect Capability \rom{1}, we further investigate whether it can benefit Capability \rom{2} by testing on distracting context. The results are visualized on Figure~\ref{fig:mixed_ft_distract_test}. As can be seen, \emph{increasing the distracting ratios steadily improves the performance when provided with distracting context}. On the easier AmbigQA, ePQA and SciQ datasets, after LLMs getting used to their input formats, the performance when provided with distracting context can be very close to the performance when no context is provided, i.e.\ , the model is not affected by the distracting context. 
This holds true for models of varying sizes, with LoRA fine-tuning performing similarly to full fine-tuning. On the more challenging datasets, Musique and TopioCQA, despite the steady improvement, there is still some room for growth before the model can be fully robust against distracting context. We hypothesize that the model may require more data to effectively understand longer input sequences, considering that Musique includes multiple context passages and TopioCQA involves an entire conversation as the input question.

In summary, when fine-tuning LLMs on a mixture of gold and distracting context, we have the following observations:
\begin{enumerate}
    \item Capability \rom{1} is maintained, or sometimes even enhanced, when the distracting ratio is increased in the fine-tuning data.
    \item Capability \rom{2} gets improved steadily. On easier datasets with shorter inputs, the model can even achieve complete robustness against distracting context.
\end{enumerate}
\section{Conclusion}
Retrieval robustness is the key to determine the quality of model generations in RAG. In this paper, we conduct an extensive assessment of the ``implicit'' retrieval robustness of LLMs without explicitly letting models judge the relevance of the retrieved context. Our findings indicate that LLMs are remarkably adept at handling context with varied retrieval accuracy, without needing explicit relevance annotations. By incorporating a certain ratio of distracting context into the fine-tuning dataset, LLMs can maintain their ability to extract correct answers from relevant context while hardly being misled by irrelevant information.
\section*{Limitations}
We aim to perform an extensive evaluation of the implicit retrieval robustness across various LLMs. However, due to resource and time constraint there are several limitations of this paper.

First, we select models based only on LLama and LLama-2 with up to 33B parameters. By the time of writing, there have been more advanced and larger open-source models available. The conclusions drawn from this paper, especially the comparison between open-source LLMs and closed-source LLMs might not hold with up-to-date models.

Second, we choose only datasets with short answers for simplicity of evaluations in this paper. Long answers are also an important research direction and is attracting growing attention. When instructing models to generate more complex long answers, the retrieval robustness of LLMs need to be re-examined.

Finally, despite conducting a grid search over a wide range of learning rates, it is possible that the optimal configuration lies outside the range we considered. We also did not extensively test results with different batch sizes and data sizes, which could impact model performance in various ways.
\section*{Ethics Statement}
Our work’s sole aim is to study the implicit retrieval robustness of retrieval-augmented large language models. We expect minimal social risks to be associated with our efforts.

\bibliography{anthology,custom}
\bibliographystyle{acl_natbib}

\newpage
\appendix

\section{Prompts used for LLMs}
\label{sec:app_prompt}

\subsection{W/o Retrieval}
\begin{promptbox}[woret1]{AmbigQA/MuSique/SciQ}
  \small \texttt{Answer the following question with less than 10 words. Question: }
  \textbf{[Q]}
\end{promptbox}

\begin{promptbox}[woret2]{ePQA}
  \small \texttt{Answer the following question about a product with less than 10 words. If it is a binary question, always begin with yes or no. Product name: }
  \textbf{[PRODUCT TITLE]} \texttt{Question: } \textbf{[QUESTION]}
\end{promptbox}

\begin{promptbox}[woret3]{TopioCQA}
  \small \texttt{Answer the following conversation with less than 10 words. Turns are split by [sep]. Conversation: }
  \textbf{[CONVERSATION]}
\end{promptbox}

\subsection{W. Retrieval}
\begin{promptbox}[wret1]{AmbigQA/MuSique/SciQ}
  \small \texttt{Answer the following question with less than 10 words. The context is retrieved information which may or may not be helpful. When the context is unhelpful, answer it with your own knowledge. Question: }\textbf{[QUESTION]} \texttt{Context: }\textbf{[CONTEXT]}
\end{promptbox}

\begin{promptbox}[wret2]{ePQA}
  \small \texttt{Answer the following question about a product with less than 10 words. If it is a binary question, always begin with yes or no. The context is retrieved information which may or may not be helpful. When the context is unhelpful, answer it with your own knowledge. Product name: }\textbf{[PRODUCT TITLE]} \texttt{Question: }\textbf{[QUESTION]} \texttt{Context: }\textbf{[CONTEXT]}
\end{promptbox}

\begin{promptbox}[wret3]{TopioCQA}
  \small \texttt{Answer the following conversation with less than 10 words. Turns are split by [sep]. The context is retrieved information which may or may not be helpful. When the context is unhelpful, answer it with your own knowledge. Conversation: }
  \textbf{[CONVERSATION]} \texttt{Context: }\textbf{[CONTEXT]}
\end{promptbox}

\section{Dataset Examples}
\label{sec:app_dataset}
\begin{table*}
\centering
\small
\begin{tabularx}{\textwidth}{ c|X|X|X }
  \toprule
  \textbf{Dataset} &\textbf{Question} & \textbf{Context} & \textbf{Answer} \\
  \midrule 
  \multirow{4}{*}{\textbf{AmbigQA}} & when did the first star wars movie come out, in less than 32 theaters?  & "star wars" debuted on wednesday, may 25, 1977, in fewer than 32 theaters, and eight more on thursday and friday. kurtz said ... & ['may 25, 1977','25th may, 1977','05/25/1977']  \\
  \midrule 
  \multirow{3}{*}{\textbf{ePQA}} & how much do
these weigh?  & item\_weight: \{ unit:ounces, normalized\_value:\{ unit:pounds, value:0.34 \}, value:5.4 \} & ['0.34 pounds',
'5.4 ounces']  \\
  \midrule 
  \multirow{4}{*}{\textbf{Musique}} & who is the father of the creator of the white rabbit?  & ["the white rabbit is a fictional character in lewis carroll's book ...", "charles dodgson was born in 1800 in hamilton ..."] & ['charles dodgson']  \\
  \midrule 
  \multirow{4}{*}{\textbf{SciQ}} & matter undergoing chemical reactions and physical
changes can release or absorb heat. a change that releases heat is called what?  & matter undergoing chemical reactions and physical changes can release or absorb heat. a
change that releases heat is called an ... & ['exothermic process']  \\
  \midrule 
  \multirow{9}{*}{\textbf{TopioCQA}} & where do guinea pigs come from in the wild [sep] they originated in the andes of south america [sep] how do they look like [sep] guinea pigs are large for rodents; the common pet breeds weigh between when full grown and measure between in length [sep] which club is associated with it & cavy clubs and associations dedicated to the showing and breeding of guinea pigs have been established worldwide. the american cavy breeders association, an adjunct to the american rabbit breeders' association, is the governing body in the united states and canada. the british cavy council ... & ['cavy clubs dedicated to the showing and breeding of guinea pigs have been established worldwide.', 'cavy clubs', 'the american cavy breeders association, british cavy council and australian national cavy council', 'cavy clubs - the american cavy breeders association']  \\
  \bottomrule
\end{tabularx}
\caption{\label{tab:dataset_examples}
\small Dataset Examples. Musique contains at least 2 passages per question as all questions require multi-hop inferences. The other datasets contain only 1 passage per question.
}
\end{table*}

Table~\ref{tab:dataset_examples} shows example snippets from each of the datasets used in this paper. Musique contains at least 2 gold passages per question as all questions require multi-hop inferences. The other datasets contain only 1 gold passage per question. When sampling distracting passages, the numper of distracting passages is the same as that of gold passages.

The original ePQA dataset contains one-sentence answers. In order to extract short answers from them, we apply ChatGPT to extract a short span from each annotated answer. If ChatGPT judges the annotated answer cannot answer the question, then we discard this example. Namely, we only keep examples that ChatGPT thinks as valid answers, so that we can reduce the chance of noisy annotations in the original dataset. For the test data, in order to catch diverse answers per question, we manually annotated other possible spans apart from the one generated by ChatGPT.

When evaluating model generations, a generation is considered correct as long as it matches any one of the gold answers. We report the maximum recall scores with all possible gold answers.

For all datasets, we select $\sim$3000 samples as the training data and 200 samples as the test data. Since our purpose is not to achieve state-of-the-art performances but rather to inspect the effects of retrieval-augmented generation, we use this data split to reduce running time.

\section{Result Tables}
\label{sec:app_result}
Table~\ref{tab:prompt},~\ref{tab:sft_gold},~\ref{tab:sft_mix80} and \ref{tab:sft_mix50} show the full results presented in this paper. We only reported the results with the best tried learning rate.

We run all experiments on 8 Nvidia A100 GPUs. Each example is cut off with 1024 sub-tokens. On each dataset, we train the model for one epoch and select the run with the best learning rate. Each training takes about 10 GPU hours for a 7B model, 15 hours for a 13B model and 30 hours for a 33B model.
\begin{table*}
\centering
\small
\begin{tabular}{l|c|ccccccc}
\toprule
\textbf{Dataset} & \textbf{Retrieval} & \textbf{Vicuna-7B} & \textbf{Vicuna-13B}  & \textbf{Vicuna-33B} & \textbf{Llama 2-7B} & \textbf{Llama 2-13B} & \textbf{GPT3.5} & \textbf{GPT4}\\
\midrule
\multirow{3}{*}{\textbf{AmbigQA}} & None& 32.75 & 41.78 & 57.59 & 37.22 & 45.78 & 57.69 & 71.07  \\
& Gold& 66.35 & 67.56 & 74.76 & 68.40& 80.95 & 85.30 & 89.98 \\ 
& Distract& 22.24 & 30.02 & 53.25 & 33.52& 41.20 & 41.93 & 52.50 \\\midrule
\multirow{3}{*}{\textbf{ePQA}} & None& 42.21 & 47.84& 49.71 & 45.17 & 45.35 & 54.83 & 55.27 \\
& Gold& 50.00 & 62.53& 63.30 & 61.44 & 51.08 & 79.78 & 77.96 \\ 
& Distract& 39.36 & 45.80 & 45.62 & 44.89 & 42.91 & 50.70 & 47.51 \\\midrule
\multirow{3}{*}{\textbf{Musique}} & None& 11.10 & 11.21 & 19.75 & 15.22 & 19.69 & 12.80 & 22.23 \\
& Gold& 39.10 & 36.60 & 43.65 & 43.63 & 40.46 & 58.56 & 74.11 \\ 
& Distract& 4.58 & 6.24 & 12.80 & 10.48 & 10.60 & 7.25 & 16.71 \\\midrule
\multirow{3}{*}{\textbf{SciQ}} & None& 45.92 & 54.75 & 61.92 & 50.75 & 53.33 & 63.67 & 71.00 \\
& Gold& 69.33 & 73.75 & 64.75 & 66.83 & 64.08 & 80.33& 90.25 \\ 
& Distract& 29.92 & 39.42 & 54.08 & 39.75 & 53.00 & 51.25& 61.50 \\\midrule
\multirow{3}{*}{\textbf{TopioCQA}} & None& 29.91 & 30.99 & 35.16 & 30.52 & 32.79 & 41.54 & 57.38 \\
& Gold& 26.87 & 30.34 & 50.19 & 34.38 & 35.56 & 64.14& 76.44 \\ 
& Distract& 19.75 & 22.13 & 28.98 & 22.65 & 21.95 & 30.87& 55.65 \\
\bottomrule
\end{tabular}
\caption{\label{tab:prompt}
Prompting Performance
}
\end{table*}

\begin{table*}
\centering
\small
\begin{tabular}{l|c|ccccccc}
\toprule
\textbf{Dataset} & \textbf{retrieval} & \textbf{Vicuna-7B} & \textbf{Vicuna-13B}  & \textbf{Vicuna-33B} & \textbf{Llama 2-7B} & \textbf{Llama 2-13B} & \textbf{GPT3.5} & \textbf{GPT4}\\
\midrule
\multicolumn{9}{c}{\textbf{full Fine-tuning}}\\
\midrule
\multirow{3}{*}{\textbf{AmbigQA}} & None&35.56&45.15&53.90&43.32&49.90 & 57.69 & 71.07  \\
& Gold&71.96&72.95&75.11&78.08&79.96 & 85.30 & 89.98 \\ 
& Distract&10.13&18.98&20.22&11.02&15.87 & 41.93 & 52.50 \\\midrule
\multirow{3}{*}{\textbf{ePQA}} & None&52.37&53.04&53.50&53.37&53.81 & 54.83 & 55.27 \\
& Gold& 79.65&79.67&81.04&78.32&81.27 & 79.78 & 77.96 \\ 
& Distract& 41.08&40.87&44.44&40.12&44.67 & 50.70 & 47.51 \\\midrule
\multirow{3}{*}{\textbf{Musique}} & None&18.30&19.77&22.82&15.40&18.75& 12.80 & 22.23 \\
& Gold&50.84&53.81&55.74&32.65&46.80 & 58.56 & 74.11 \\ 
& Distract&5.02&4.59&4.55&3.96&9.81 & 7.25 & 16.71 \\\midrule
\multirow{3}{*}{\textbf{SciQ}} & None&56.16&57.58&64.25&58.25&60.25 & 63.67 & 71.00 \\
& Gold& 81.42&83.75&82.00&82.75&76.25 & 80.33& 90.25 \\ 
& Distract&21.25&22.58&38.00&32.00&38.50 & 51.25& 61.50 \\\midrule
\multirow{3}{*}{\textbf{TopioCQA}} & None&43.44&45.39&48.58&42.77&44.59& 41.54 & 57.38 \\
& Gold&65.59&65.78&71.37&70.81&69.94 & 64.14& 76.44 \\ 
& Distract&19.77&23.79&24.82&20.16&25.21 & 30.87& 55.65 \\
\midrule
\multicolumn{9}{c}{\textbf{LoRA Fine-tuning}}\\
\midrule
\multirow{3}{*}{\textbf{AmbigQA}} & None&32.76&46.46&55.55&41.72&47.13& 57.69 & 71.07  \\
& Gold&71.49&73.52&74.18&78.70&79.51 & 85.30 & 89.98 \\ 
& Distract&10.34&13.42&53.25&8.35&40.47 & 41.93 & 52.50 \\\midrule
\multirow{3}{*}{\textbf{ePQA}} & None&52.62&53.75&54.37&49.04&53.74& 54.83 & 55.27 \\
& Gold&77.52&80.97&83.96&74.88&79.60& 79.78 & 77.96 \\ 
& Distract&43.82&41.09&43.19&41.12&47.82& 50.70 & 47.51 \\\midrule
\multirow{3}{*}{\textbf{Musique}} & None&16.06&21.54&23.58&14.62&19.18& 12.80 & 22.23 \\
& Gold&50.08&55.91&56.16&44.13&40.69 & 58.56 & 74.11 \\ 
& Distract&4.89&5.20&5.65&10.56&10.60 & 7.25 & 16.71 \\\midrule
\multirow{3}{*}{\textbf{SciQ}} & None&57.08&59.91&64.25&57.99&62.25 & 63.67 & 71.00 \\
& Gold&82.33&85.42&85.17&83.42&82.67 & 80.33& 90.25 \\ 
& Distract&23.83&29.58&46.00&31.00&35.25 & 51.25& 61.50 \\\midrule
\multirow{3}{*}{\textbf{TopioCQA}} & None&43.00&42.67&48.14&41.19&44.33& 41.54 & 57.38 \\
& Gold&64.18&63.82&68.80&68.37&68.83 & 64.14& 76.44 \\ 
& Distract&21.63&24.13&23.02&24.61&25.89 & 30.87 & 55.65 \\
\bottomrule
\end{tabular}
\caption{\label{tab:sft_gold}
Performance by Fine-Tuning on Gold retrieval
}
\end{table*}

\begin{table*}
\centering
\small
\begin{tabular}{l|c|ccccccc}
\toprule
\textbf{Dataset} & \textbf{retrieval} & \textbf{Vicuna-7B} & \textbf{Vicuna-13B}  & \textbf{Vicuna-33B} & \textbf{Llama 2-7B} & \textbf{Llama 2-13B} & \textbf{GPT3.5} & \textbf{GPT4}\\
\midrule
\multicolumn{9}{c}{\textbf{full Fine-tuning}}\\
\midrule
\multirow{3}{*}{\textbf{AmbigQA}} & None&35.56&45.15&53.90&43.32&49.90 & 57.69 & 71.07  \\
& Gold&74.62&74.36&74.10&79.94&77.24 & 85.30 & 89.98 \\ 
& Distract&30.56&31.60&51.32&34.37&43.25 & 41.93 & 52.50 \\\midrule
\multirow{3}{*}{\textbf{ePQA}} & None&52.37&53.04&53.50&53.37&53.81 & 54.83 & 55.27 \\
& Gold&75.72&79.43&81.17&77.17&81.97 & 79.78 & 77.96 \\ 
& Distract&46.91&50.73&52.69&44.66&52.69 & 50.70 & 47.51 \\\midrule
\multirow{3}{*}{\textbf{Musique}} & None&18.30&19.77&22.82&15.40&18.75& 12.80 & 22.23 \\
& Gold&49.91&53.79&57.26&50.80&55.03 & 58.56 & 74.11 \\ 
& Distract&7.26&11.73&10.66&10.54&16.43& 7.25 & 16.71 \\\midrule
\multirow{3}{*}{\textbf{SciQ}} & None&56.16&57.58&64.25&58.25&60.25 & 63.67 & 71.00 \\
& Gold&79.50&86.00&82.08&81.08&83.58& 80.33& 90.25 \\ 
& Distract&50.50&53.83&58.42&50.83&53.08 & 51.25& 61.50 \\\midrule
\multirow{3}{*}{\textbf{TopioCQA}} & None&43.44&45.39&48.58&42.77&44.59& 41.54 & 57.38 \\
& Gold&67.17&63.75&68.28&69.87&69.80 & 64.14& 76.44 \\ 
& Distract&27.35&33.22&41.91&36.41&39.92 & 30.87 & 55.65 \\
\midrule
\multicolumn{9}{c}{\textbf{LoRA Fine-tuning}}\\
\midrule
\multirow{3}{*}{\textbf{AmbigQA}} & None&32.76&46.46&55.55&41.72&47.13& 57.69 & 71.07  \\
& Gold&72.22&73.21&77.80&76.68&80.45& 85.30 & 89.98 \\ 
& Distract&29.01&38.79&43.90&33.30&41.69 & 41.93 & 52.50 \\\midrule
\multirow{3}{*}{\textbf{ePQA}} & None&52.62&53.75&54.37&49.04&53.74& 54.83 & 55.27 \\
& Gold&80.15&80.34&81.89&73.39&81.20& 79.78 & 77.96 \\ 
& Distract&45.66&50.91&49.19&51.85&52.86& 50.70 & 47.51 \\\midrule
\multirow{3}{*}{\textbf{Musique}} & None&16.06&21.54&23.58&14.62&19.18& 12.80 & 22.23 \\
& Gold&52.11&53.49&54.76&51.51&54.17& 58.56 & 74.11 \\ 
& Distract&7.97&8.35&12.32&7.37&11.03 & 7.25 & 16.71 \\\midrule
\multirow{3}{*}{\textbf{SciQ}} & None&57.08&59.91&64.25&57.99&62.25 & 63.67 & 71.00 \\
& Gold&81.75&84.67&83.75&82.25&83.42 & 80.33& 90.25 \\ 
& Distract&51.08&54.92&59.25&48.92&54.33 & 51.25& 61.50 \\\midrule
\multirow{3}{*}{\textbf{TopioCQA}} & None&43.00&42.67&48.14&41.19&44.33& 41.54 & 57.38 \\
& Gold&64.96&62.97&66.46&68.27&68.31 & 64.14& 76.44 \\ 
& Distract&30.55&32.29&41.11&30.63&37.80&30.87&55.65 \\
\bottomrule
\end{tabular}
\caption{\label{tab:sft_mix80}
Performance by Fine-Tuning on 80\% Gold + 20\% Distracting retrieval
}
\end{table*}

\begin{table*}
\centering
\small
\begin{tabular}{l|c|ccccccc}
\toprule
\textbf{Dataset} & \textbf{retrieval} & \textbf{Vicuna-7B} & \textbf{Vicuna-13B}  & \textbf{Vicuna-33B} & \textbf{Llama 2-7B} & \textbf{Llama 2-13B} & \textbf{GPT3.5} & \textbf{GPT4}\\
\midrule
\multicolumn{9}{c}{\textbf{full Fine-tuning}}\\
\midrule
\multirow{3}{*}{\textbf{AmbigQA}} & None&35.56&45.15&53.90&43.32&49.90 & 57.69 & 71.07  \\
& Gold&72.18&74.33&74.55&75.98&76.77 & 85.30 & 89.98 \\ 
& Distract&31.16&44.10&53.11&36.34&43.05& 41.93 & 52.50 \\\midrule
\multirow{3}{*}{\textbf{ePQA}} & None&52.37&53.04&53.50&53.37&53.81 & 54.83 & 55.27 \\
& Gold&77.59&77.92&82.47&77.95&79.14 & 79.78 & 77.96 \\ 
& Distract&51.87&52.95&53.33&51.45&53.95 & 50.70 & 47.51 \\\midrule
\multirow{3}{*}{\textbf{Musique}} & None&18.30&19.77&22.82&15.40&18.75& 12.80 & 22.23 \\
& Gold&49.39&48.86&51.37&51.40&52.25 & 58.56 & 74.11 \\ 
& Distract&12.26&12.03&15.28&8.11&14.68& 7.25 & 16.71 \\\midrule
\multirow{3}{*}{\textbf{SciQ}} & None&56.16&57.58&64.25&58.25&60.25 & 63.67 & 71.00 \\
& Gold&80.75&81.33&80.25&80.17&84.00& 80.33& 90.25 \\ 
& Distract&51.67&56.99&61.91&50.92&53.58 & 51.25& 61.50 \\\midrule
\multirow{3}{*}{\textbf{TopioCQA}} & None&43.44&45.39&48.58&42.77&44.59& 41.54 & 57.38 \\
& Gold&67.13&64.73&68.08&70.00&68.96 & 64.14& 76.44 \\ 
& Distract&32.57&35.61&44.60&37.97&39.42 & 30.87& 55.65 \\
\midrule
\multicolumn{9}{c}{\textbf{LoRA Fine-tuning}}\\
\midrule
\multirow{3}{*}{\textbf{AmbigQA}} & None&32.76&46.46&55.55&41.72&47.13& 57.69 & 71.07  \\
& Gold&74.22&73.32&74.18&73.02&79.34& 85.30 & 89.98 \\ 
& Distract&33.04&42.80&52.75&35.26&40.47 & 41.93 & 52.50 \\\midrule
\multirow{3}{*}{\textbf{ePQA}} & None&52.62&53.75&54.37&49.04&53.74& 54.83 & 55.27 \\
& Gold&77.59&77.72&80.94&73.76&81.42& 79.78 & 77.96 \\ 
& Distract&51.16&50.07&52.44&53.44&54.29& 50.70 & 47.51 \\\midrule
\multirow{3}{*}{\textbf{Musique}} & None&16.06&21.54&23.58&14.62&19.18& 12.80 & 22.23 \\
& Gold&49.03&49.30&51.71&47.65&49.84& 58.56 & 74.11 \\ 
& Distract&9.33&11.36&14.74&10.38&12.76 & 7.25 & 16.71 \\\midrule
\multirow{3}{*}{\textbf{SciQ}} & None&57.08&59.91&64.25&57.99&62.25 & 63.67 & 71.00 \\
& Gold&80.08&84.08&83.33&82.67&82.58 & 80.33& 90.25 \\ 
& Distract&52.50&53.75&62.00&50.58&56.16 & 51.25& 61.50 \\\midrule
\multirow{3}{*}{\textbf{TopioCQA}} & None&43.00&42.67&48.14&41.19&44.33& 41.54 & 57.38 \\
& Gold&67.14&61.62&67.74&65.06&68.57& 64.14& 76.44 \\ 
& Distract&34.67&33.54&40.83&38.56&40.94&30.87&55.65 \\
\bottomrule
\end{tabular}
\caption{\label{tab:sft_mix50}
Performance by Fine-Tuning on 50\% Gold + 50\% Distracting retrieval
}
\end{table*}
\end{document}